\def\year{2017}\relax
\documentclass[letterpaper]{article}
\usepackage{aaai17}
\usepackage{times}
\usepackage{helvet}
\usepackage{courier}
\usepackage{color}
\usepackage{amsmath}
\usepackage{multirow}
\usepackage{url}
\usepackage{graphicx}
\usepackage{subfigure}
\usepackage{paralist}

\begin{document}
%
\title{Language Use Matters: Analysis of the Linguistic Structure of Question Texts Can Characterize Answerability in Quora}
\author{Suman Kalyan Maity\textsuperscript{1}, Aman Kharb\textsuperscript{2} and Animesh Mukherjee\textsuperscript{3}\\
Department of Computer Science and Engineering\\
Indian Institute of Technology Kharagpur, India - 721302\\
Email:  \{sumankalyan.maity\textsuperscript{1},animeshm\textsuperscript{3}\}@cse.iitkgp.ernet.in; manu2kharb@gmail.com\textsuperscript{2} \\
}

\maketitle
\begin{abstract}
Quora is one of the most popular community Q\&A sites of recent times. However, many question posts on this Q\&A site often do not get answered. In this paper, we quantify various linguistic activities that discriminates an answered question from an unanswered one. Our central finding is that the way users use language while writing the question text can be a very effective means to characterize answerability. This characterization helps us to predict early if a question remaining unanswered for a specific time period $t$ will eventually be answered or not and achieve an accuracy of {\bf 76.26}\% ($t=1$ month) and {\bf 68.33}\% ($t=3$ months). Notably, features representing the language use patterns of the users are most discriminative and alone account for an accuracy of {\bf 74.18}\%. We also compare our method with some of the similar works (Dror et al., Yang et al.) achieving a maximum improvement of $\sim39\%$ in terms of accuracy.
\end{abstract}

\section{Introduction}
From a group of small users at the time of its inception in 2009, Quora has evolved in the last few years into one of the largest community driven Q\&A sites with diverse user communities. With the help of efficient content moderation/review policies and active in-house review team, efficient Quora bots, this site has emerged into one of the largest and reliable sources of Q\&A on the Internet. On Quora, users can post questions, follow questions, share questions, tag them with relevant topics, follow topics, follow users apart from answering, commenting, upvoting/downvoting etc. The integrated social structure at the backbone of it and the topical organization of its rich content have made Quora unique with respect to other Q\&A sites like Stack Overflow, Yahoo! Answers etc. and these are some of the prime reasons behind its popularity in recent times. Quality question posting and getting them answered are the key objectives of any Q\&A site. In this study we focus on the answerability of questions on Quora, i.e., whether a posted question shall eventually get answered. In Quora, the questions with no answers are referred to as ``open questions''. These open questions need to be studied separately to understand the reason behind their not being answered or to be precise, are there any characteristic differences between `open' questions and the answered ones. For example, the question ``What are the most promising advances in the treatment of traumatic brain injuries?'' was posted on Quora on $23^{rd}$ June, 2011 and got its first answer after almost 2 years on $22^{nd}$ April, 2013. The reason that this question remained open so long might be the hardness of answering it and the lack of visibility and experts in the domain. Therefore, it is important to identify the open questions and take measures based on the types - poor quality questions can be removed from Quora and the good quality questions can be promoted so that they get more visibility and are eventually routed to topical experts for better answers. 

Characterization of the questions based on question quality requires expert human interventions often judging if a question would remain open based on factors like if it is subjective, controversial, open-ended, vague/imprecise, ill-formed, off-topic, ambiguous, uninteresting etc. Collecting judgment data for thousands of question posts is a very expensive process. Therefore, such an experiment can be done only for a small set of questions and it would be practically impossible to scale it up for the entire collection of posts on the Q\&A site. In this work, we show that appropriate quantification of various {\em linguistic} activities can naturally correspond to many of the judgment factors mentioned above (see table~\ref{tab:examples} for a collection of examples). These quantities encoding such linguistic activities can be easily measured for each question post and thus helps us to have an alternative mechanism to characterize the answerability on the Q\&A site.

There are several research works done in Q\&A focusing on content of posts. ~\citeauthor{agi} exploit community feedback to identify high quality content on Yahoo! Answers.~\citeauthor{shah} use textual features to predict answer quality on Yahoo! Answers.~\citeauthor{maxwell}, investigate predictors of answer quality through a comparative, controlled field study of user responses. ~\citeauthor{asad} study the problem of how long questions remain unanswered.~\citeauthor{dror} propose a prediction model on how many answers a question shall receive.~\citeauthor{Yang:2011} analyze and predict unanswered questions on Yahoo Answers.~\citeauthor{li} study question quality in Yahoo! Answers. 

\section{Dataset description}\label{data}
We obtained our Quora dataset~\cite{suman} through web-based crawls between June 2014 to August 2014. This crawling exercise has resulted in the accumulation of a massive Q\&A dataset spanning over a period of over four years starting from January 2010 to May 2014. We initiated crawling with 100 questions randomly selected from different topics so that different genre of questions can be covered. The crawling of the questions follow a BFS pattern through the related question links. We obtained 822,040 unique questions across 80,253 different topics with a total of 1,833,125 answers to these questions. For each question, we separately crawl their revision logs that contain different types of edit information for the question and the activity log of the question asker.
\section{Linguistic activities on Quora}\label{linguistic}
In this section, we identify various linguistic activities on Quora and propose quantifications of the language usage patterns in this Q\&A site. In particular, we show that there exists significant differences in the linguistic structure of the open and the answered questions. Note that most of the measures that we define are simple, intuitive and can be easily obtained automatically from the data (without manual intervention). Therefore the framework is practical, inexpensive and highly scalable. 

Content of a question text is important to attract people and make them engage more toward it. The linguistic structure (i.e., the usage of POS tags, the use of Out-of-Vocabulary words, character usage etc.) one adopts are key factors for answerability of questions. We shall discuss the linguistic structure that often represents the writing style of a question asker.

\begin{figure}[ht]
\begin{center}
\includegraphics*[scale=0.25,angle=0]{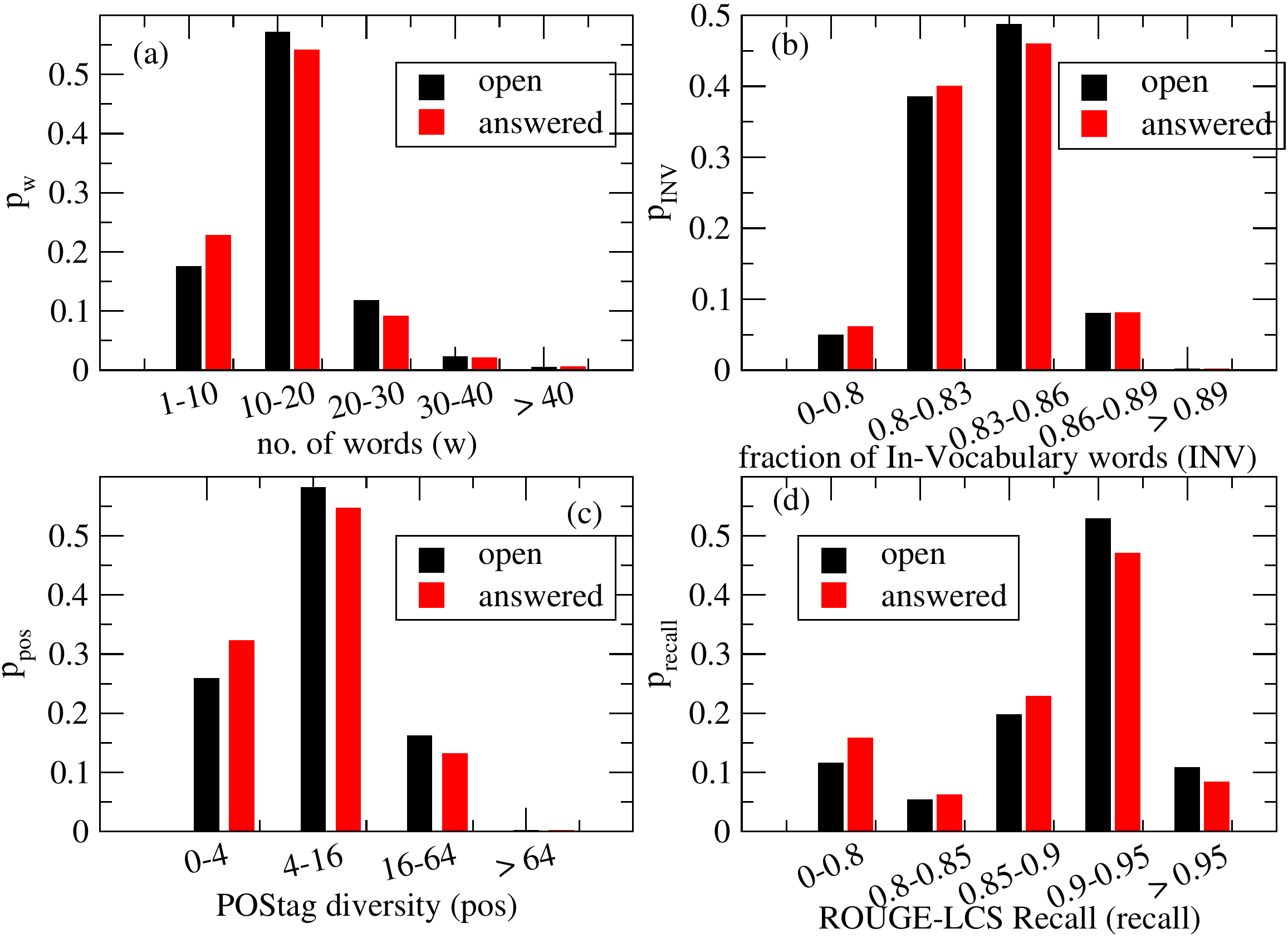}
\caption{\label{figlinguistic} Comparison of distribution of a) no. of words in the question b) fraction of In-Vocabulary words c) POS tag diversity d) ROUGE-LCS recall for open questions vs answered question.
}
\end{center}
\end{figure}

In fig~\ref{figlinguistic}(a), we observe that askers of open questions generally use more no. of words compared to answered questions. To understand the nature of words (standard English words or chat-like words frequently used in social media) used in the text, we compare the words with GNU Aspell dictionary\footnote{\label{oov}\url{http://aspell.net/}} to see whether they are present in the dictionary or not. We observe that both open questions and answered questions follow similar distribution (see fig~\ref{figlinguistic}(b)). Part-of-Speech (POS) tags are indicators of grammatical aspects of texts. To observe how the Part-of-Speech tags are distributed in the question texts, we define a diversity metric. We use the standard CMU POS tagger~\cite{owu} for identifying the POS tags of the constituent words in the question. We define the POS tag diversity (POSDiv) of a question $q_i$ as follows: $POSDiv(q_i) = -\sum_{j \in pos_{set}}p_j\times\log(p_j)$ 
 where $p_j$ is the probability of the $j^{th}$ POS in the set of POS tags. Fig~\ref{figlinguistic}(c) shows that the answered questions have lower POS tag diversity compared to open questions. Question texts undergo several edits so that their readability and the engagement toward them are enhanced. It is interesting to identify how far such edits can make the question different from the original version of it. To capture this phenomena, we have adopted ROUGE-LCS recall~\cite{rouge} from the domain of text summarization. Higher the recall value, lesser are the changes in the question text. From fig~\ref{figlinguistic}(d), we observe that open questions tend to have higher recall compared to the answered ones which suggests that they have not gone through much of text editing thus allowing for almost no scope of readability enhancement. 

\noindent \textbf{Psycholinguistic analysis: }
\begin{table}
\caption{LIWC analysis for open and answered questions.}
\label{tab:liwc}
\centering
\resizebox{7.4cm}{!}{
\begin{tabular}{|p{5cm}| p{3.5cm}| p{3.5cm}|}
\hline\
LIWC category & Avg. LIWC score for open questions & Avg. LIWC score for answered questions \\ \hline
\multicolumn{3}{|c|}{Linguistic processes} \\\hline
Function words & 53.4103535493 & 50.6851839369\\ 
Pronouns & 12.465131081 & 8.9026143697\\ 
Personal pronouns & 2.0535742638 & 3.2504745366\\ 
1st person singular & 0.5472352995 & 1.25078055\\ 
1st person plural & 0.2264298902 & 0.3101397509\\ 
2nd person & 0.8891512047 & 0.9988166386\\ 
3rd person singular & 0.0872251454 & 0.1513366764\\ 
3rd person plural & 0.3009577218 & 0.538239755\\ 
Impersonal pronoun & 10.41145066 & 5.652067286\\ 
Articles & 9.5123765347 & 6.8941214189\\ 
Adverbs & 1.8814285055 & 4.7420948251\\ 
Conjunctions & 2.9557966399 & 5.5545170373\\ 
Negation & 0.2122514539 & 0.5633840273\\ \hline
\multicolumn{3}{|c|}{Psychological processes} \\\hline
Social process & 5.1287708853 & 5.7786061901\\ 
Friends & 0.0862088064 & 0.113930289\\ 
Humans & 0.6565905105 & 0.7599064902\\ 
Positive Emotion & 4.6592739777 & 3.237412964\\ 
Negative Emotion & 0.6840035078 & 0.8363457191\\ 
Anxiety & 0.0840210468 & 0.1076020387\\ 
Anger & 0.2170340626 & 0.2476001891\\ 
Sadness & 0.1204472445 & 0.1710861523\\ 
Cognitive Processes & 10.0861552663 & 15.3887239508\\ 
Cause & 1.8814640451 & 5.1099404004\\ 
Tentative & 1.997843626 & 3.4358232973\\ 
Biological Processes & 1.148483338 & 1.111548769\\ 
Body & 0.266928367 & 0.2620352255\\ 
Health & 0.4831117881 & 0.477567512\\ 
Sexual & 0.0809014124 & 0.0779684738\\ \hline
\end{tabular}}
\end{table}
The way an individual talks or writes, give us clue to his/her linguistic, emotional, and cognitive states. A question asker's linguistic, emotional, cognitive states are also revealed through the language he/she uses in the question text. In order to capture such psycholinguistic aspects of the asker, we use Linguistic Inquiry and Word Count (LIWC)~\cite{liwc} that analyzes various emotional, cognitive, and structural components present in individuals' written texts. LIWC takes a text document as input and outputs a score for the input for each of the LIWC categories such as linguistic (part-of-speech of the words, function words etc.) and psychological categories (social, anger, positive emotion, negative emotion, sadness etc.) based on the writing style and psychometric properties of the document. In table~\ref{tab:liwc}, we perform a comparative analysis of the asker's psycholinguistic state while asking an open question and an answered question.

Askers of open questions use more function words, impersonal pronouns, articles on an average whereas asker of answered questions use more personal pronouns, conjunctions and adverbs to describe their questions. Essentially, open questions lack content words compared to answered questions which, in turn, affects the readability of the question. As far as the psychological aspects are concerned, answered question askers tend to use more social, family, human related words on average compared to an open question asker. The open question askers express more positive emotions whereas the answered question asker tend to express more negative emotions in their texts. Also, answered question askers are more emotionally involved and their questions reveal higher usage of anger, sadness, anxiety related words compared to that of open questions. Open questions, on the other hand, contains more sexual, body, health related words which might be reasons why they do not attract answers.

In table~\ref{tab:examples}, we show a collection of examples of open questions to illustrate that many of the above quantities based on the linguistic activities described in this section naturally correspond to the factors that human judges consider responsible for a question remaining unanswered. This is one of the prime reasons why these quantities qualify as appropriate indicators of answerability. 

\begin{table*}
\caption{Examples of open questions w.r.t. various linguistic activities}
\label{tab:examples}
\centering
\resizebox{15cm}{!}{
\begin{tabular}{|p{15cm}| p{4cm} | p{4cm} |} 
\hline\
Open questions & Linguistic activities & Characteristics \\ \hline
Why Is Facebook And The Ad Agencies That Make Money Off Facebook Blaming Their Very Clients For The Poor Results Experienced On Facebook's Platform?  Is That Just A Subterfuge? & high POS tag diversity, lengthy & too controversial, infuses debates/discussions \\\hline
How does Max Weinberg feel to be the only person to be both on the last episode of Late Night with David Letterman (as the drummer for guest Bruce Springsteen) and the first episode of Late Night with Conan O'Brien (as the house bandleader)? & high POS tag diversity, lengthy & ill-formed, not specific, vague, too many queries jumbled up \\\hline
If a warehouse of physical goods is seized in the US because of illegal activity by the owner and a few customers using it, are the authorities required to return items that are "innocent" and were collateral damage once the investigation is complete? & high POS tag diversity, high ROUGE-LCS score, lengthy & vague, ill-explained, requires experts to answer \\\hline
How can Matthew Reilly write such astounding action books? How does he prepare himself while writing a new novel? & low ROUGE-LCS score & Very opinionated, difficult to answer \\\hline
1) How expensive it is to get into Big Data Analytics area with simple service offerings? 2) What is the most simple and popular service provided by companies? I would appreciate an early response on the above or pointers to knowledge sources. & lengthy, high POS tag diversity& too many questions, vague/imprecise \\\hline

\end{tabular}}
\end{table*}
\section{Prediction model}\label{pred}
In this section, we describe the prediction framework in detail. Our goal is to predict whether a given question after a time period $t$ will be answered or not. \\
\textbf{Linguistic styles of the question asker} \\
The content and way of posing a question is important to attract answers. We have observed in the previous section that these linguistic as well as psycholinguistic aspects of the question asker are discriminatory factors. For the prediction, we use the following features:
\begin{compactitem}
\item Character length of a question, number of words in a question, fraction of non-frequent words in a question, and number of function words in a question.
 \item In-Vocabulary (INV) words and Out-of-Vocabulary (OOV) words in question text - we check whether a word appearing in the question text, is an INV or OOV word by comparing with GNU Aspell dictionary. We then consider the fraction of INV words as a feature of our model. 
 \item Presence of n-grams of the question content in English texts - we search for 2, 3, 4 grams of the words from the question text in the corpus of 1 million contemporary American English words\footnote{\url{http://www.ngrams.info/samples\_coca1.asp}}. We use the presence of bigrams, trigrams, 4-grams each as features.
 \item POS tag diversity - we use the difference in POS tag diversity between the initial question text and the question text after time period $t$ (observation period) as a feature.
\item Distribution of LDA topics obtained from question texts - for topic discovery from the question corpus, we adopt the popular LDA model~\cite{lda}. For a question $q_i$, we consider all the words in that question as a document. We set the number of topics as K = 10, 20, 30 and find out $p(topic_k|D_i)$ for a document $D_i$ containing all the words of the $i^{th}$ question. Each of these $p(topic_k|D_i)$ for $k = 1..K$ act as a feature.
\item LDA topical diversity - we also compute LDA topical diversity ($TopicDiv$) of a question ($q_i$) from the document-topic distributions obtained above as follows and use this metric as a feature. We define $TopicDiv(q_i) = -\sum_{k=1}^{K}p(topic_k|D_i) \times\log p(topic_k|D_i)$.
\item Psycholinguistic aspects of question texts - we consider the LIWC scores from the different categories as features for the model.
\item ROUGE-LCS recall of the question text at the end of the observation period of the prediction with reference to the original question text posted by the asker.
\end{compactitem}

\noindent\textbf{Question editing activities}: We also consider various editing activities in questions which could also be a potential source of difference.
\begin{compactitem}
\item Number of (i) context topic edits, (ii) question text edits, (iii) question detail edits, (iv) times new topics have been added to the question, (v) times existing topics have been removed from the question, (vi) times topics added by users other than the asker, (vii) topic edits done by the Quora review team, and (viii) other kinds of edits done by the Quora review team.
\item Question promotions - A question can be promoted to various users for increased visibility. We use number of question promotions as well as the number of people to which it has been promoted as features.
\item Average time interval between edits.
\end{compactitem}
\textbf{Features based on topic hierarchy}: Question topics play an important role in organizing the question and better the organization a question has, better is its chance of exposure to the experts. In Quora, topics are hierarchically organized via parent-child relationship in the form of forests with a core tree. We separately crawl the topic hierarchy of almost all the topics available in Quora and devise the following features  
\begin{compactitem}
 \item Number of topics associated with a question.
 \item Average depth, maximum depth and variance of depth of the question topics in the topic hierarchy.
 \item Maximum number of question topics belonging to the same level in the topic hierarchy tree.
 \item Number of connected components of the topic hierarchy graph the question topics belong to.
 \item Difference in question topics at the time of question post and the topics associated with the question after time period $t$ (observation period).
\end{compactitem}
\section{Performance of the prediction model}\label{eval} 
We perform our predictions at two time points -- $t = 1$ month and $t = 3$ months after a question is posted. In other words, for the first (second) case any question that remains open at the end of one month (three months) is labeled as `open' in the ground-truth data, else it is labeled `answered'. Further, in the first (second) case, all the features described in the previous section are calculated only using the one month (three months) observation data. Restricting the computation of the features to the observation period only ensures that there is strictly no scope for data leakage. 

In the prediction task, we remove all the questions posted by the anonymous users. We perform a 10-fold cross-validation with SVM classifier and achieve {\bf 76.26}\% accuracy with high avg. precision and recall rates for $t$ = 1 month and {\bf 68.33}\% for $t$ = 3 months (see table~\ref{class} for details). Logistic regression (LR) and random forest (RF) classifiers yield very similar classification performance (at $t = 1$ month, accuracy of {\bf 75.11}\% and {\bf 74.42}\% for LR and RF respectively) although SVM performs best among them. Consequently, we report the performance of the SVM classifier in detail. We observe that the number of topics ($K$) of LDA does not have a significant effect on the classification results. For $K = 20$, we achieve the best accuracy, avg. precision, recall and the area under the ROC curve. Note that, as time progresses prediction becomes harder; thanks to the rich set of features, even with three months observation period, we are able to obtain a decent prediction accuracy. 

There are a very few early works~\cite{Yang:2011,dror} regarding answerability of the questions and we use them as baseline models. We achieve $\sim33\%$ and $\sim39\%$ improvement in terms of accuracy over Yang et al.'s and Dror et al.'s method respectively for $t = 1$ month (see table~\ref{class} for further details). Our method is performing best for prediction on shorter time periods than the baselines.
\begin{table}[h]
\centering
\caption{Performance of various methods ($K$ = 10, 20, 30). First 5 lines for $t$ = 1 month and last 5 lines for $t$ = 3 months}
\label{class}
  \resizebox{8cm}{!}{
 \begin{tabular}{ |c|c|c|c|c|c|c| }
\hline
Method & $K$ & Accu\-racy &Preci\-sion & Recall & F-Score & ROC Area \\ \hline
 \multirow{3}{2cm}{Our Method} & 10 & 75.21\% & 0.752 & 0.752 & 0.752 &0.752 \\\cline{2-7}
 & 20 & \textbf{76.26\%} & 0.763 & 0.763 & 0.763 &  0.762 \\ \cline{2-7}
 & 30 & 76.11\% & 0.761 & 0.761 & 0.761 &  0.761 \\ \cline{1-7}
 Yang et al.&  & 57.4\%& 0.534 & 0.734 & 0.618 & 0.543\\ \cline{1-7}
 Dror et al. & & 55\% & 0.543 &0.73 & 0.624 & 0.554 \\ \hline \hline
\multirow{3}{2cm}{Our Method } & 10 & 64.3\% & 0.643 &  0.643 &  0.643 & 0.643 \\ \cline{2-7}
  & 20 & \textbf{68.33\%} & 0.684 & 0.683 & 0.683 &  0.683 \\ \cline{2-7}
 & 30 & 66.8\% & 0.669 & 0.669 & 0.669 & 0.669 \\ \cline{1-7}
Yang et al. & &59.3\%  & 0.587 & 0.64 & 0.613 & 0.592 \\ \cline{1-7}
Dror et al. & & 59.8\% & 0.596 & 0.628& 0.612 & 0.598 \\ \hline
\end{tabular}}
\end{table} \\
\noindent\textbf{Feature importance:} We observe that linguistic style features are the most discriminative ones achieving an accuracy of {\bf 74.18}\% alone. In order to further determine the discriminative power of each feature, we compute the $\chi^2$ value and the information gain. The most prominent ones among the linguistic activities are the LIWC features. Some other features that are effective are the topic hierarchy and the topical edit features.
\section{Conclusions}\label{conc}
In this paper, we investigate various linguistic activities and observe how such activities affect answerability of questions in Quora. One of the primary lesson is that the language usage patterns correspond to quality factors that human judges would consider to decide if a question would remain unanswered. Based on these linguistic activities we can efficiently discriminate the open and the answered questions. Our proposed prediction framework achieves an accuracy of {\bf 76.26}\% ($t=1$ month) and {\bf 68.33}\% ($t=1$ months) with high precision and recall outperforming the baseline methods convincingly.

\fontsize{9.0pt}{10.0pt}
\selectfont
\bibliography{ref}
\bibliographystyle{aaai}
\end{document}